\documentclass[10pt,A4,twoside]{article}

\usepackage{multirow}
\usepackage{graphicx}
\usepackage{times}
\usepackage[applemac]{inputenc}
\usepackage[T1]{fontenc}

\usepackage{taln2011}
\usepackage[frenchb]{babel}

\title{Une analyse basée sur la S-DRT pour la modélisation de dialogues pathologiques}

\author{Maxime Amblard \up{1,4}\quad Michel Musiol\up{2,4} \quad Manuel Rebuschi\up{3,4}\\
  (1) LORIA - UMR 7503 \\ 
  (2) InterPSY - EA 4432 / MSH Lorraine USR 3261\\ 
  (3) Archives Poincaré - UMR 7117 / MSH Lorraine USR 3261\\ 
  (4) Université Nancy 2 -- 54000 Nancy \\ 
  \{maxime.amblard, michel.musiol, manuel.rebuschi\}@univ-nancy2.fr\\ 
}

\begin{document}

\maketitle

\resume{Dans cet article, nous présentons la définition et l'étude d'un corpus de dialogues entre un schizophrène et un interlocuteur ayant pour objectif la conduite et le maintien de l'échange. Nous avons identifié des discontinuités significatives chez les schizophrènes paranoïdes. Une représentation issue de la S-DRT (sa partie pragmatique) permet de rendre compte des ces usages non standards.
}

\abstract{In this article, we present a corpus of dialogues between a schizophrenic speaker and an interlocutor who drives the dialogue. We had identified specific discontinuities for paranoid schizophrenics. We propose a modeling of these discontinuities with S-DRT (its pragmatic part).
}

\motsClefs{S-DRT, interaction verbale, schizophrénie, dialogue pathologique, incohérence pragmatique}
{S-DRT, verbal interaction, schizophrenia, pathological dialogue, pragmatical incoherence }

\section{Contexte Scientifique}
La pathologie schizophrénique constitue aujourd’hui encore une entité clinique complexe et mal définie. Elle suscite de nombreuses controverses relativement aux caractéristiques symptomatologiques (ou regroupements syndromiques) susceptibles de la définir. Il est difficile de trouver des caractéristiques ou traits partagés par les individus présentant ce diagnostic et nous ne disposons d’aucun signe pathognomonique clairement défini dans la littérature scientifique qui puisse la spécifier. Ce constat théorique est relayé par la multiplicité des manifestations cliniques présentées. Quant à la manifestation objective des symptômes, il est encore impossible de rapporter sans risque des traits de comportements manifestes et identifiables à des caractéristiques syndromiques circonscrites.

Nous estimons que les productions et manifestations comportementales de tout sujet, "normal" ou pathologique, sont nécessairement soumises à l’épreuve d’un cadre interactionnel et discursif, fût-il expérimental ou clinique. Nous formulons l’hypothèse selon laquelle le comportement verbal de tout (inter)-locuteur est susceptible de refléter des spécificités syndromiques ; ce comportement s’étaye sur un ensemble de contraintes sociales et cognitives qui sont la condition de l’usage naturel de la langue. L’interaction verbale est alors considérée comme le "lieu naturel d’expression des symptômes" \cite{MusiolTrognon96}. Nous envisageons de mettre au jour le plus objectivement possible, c’est-à-dire de manière "décisive"\footnote{Cf plus bas. Les séquences conversationnelles à ruptures décisives ne résistent pas à l’épreuve du principe logique de non-contradiction. Les séquences conversationnelles à ruptures non décisives présentent des caractéristiques incongrues ou des formes d’incohérences étayant des infractions comportementales de type "normatives".}, les discontinuités apparaissant dans le discours et le dialogue, discontinuités dont on discutera ensuite la relation à de possibles spécificités syndromiques. Ce programme de recherche a aussi pour ambition de traiter de l’interprétation de ces discontinuités en termes d’incohérences ou de dysfonctionnements. Sur le plan méthodologique, nous nous proposons de spécifier la notion d’incohérence en types ou modèles de discontinuité de l’interaction verbale. Et nous proposons plus globalement un programme d’analyse de l’interaction verbale d’inspiration pragmatique, cognitive et formelle \cite{MusiolRebuschi07, RebMus10, AmMuRe10} tel que le repérage et la description de ces spécificités (ou discontinuités), quand il y en a, devrait améliorer à moyen terme à la fois les stratégies de diagnostic usuelles et les tentatives de spécification des troubles sur le plan des opérations cognitives et de pensée complexes. Les modalités d’expression du trouble sont appréhendées dans la structure intentionnelle du dialogue. Ainsi, à condition de disposer d’une stratégie de modélisation et de formalisation adéquate, cette structure intentionnelle du dialogue nous livre en filigrane les principales propriétés de la rationalité du trouble dans sa modalité inférentielle et computo-représentationnelle.

Le programme de recherche d’inspiration pragmatique, dialogique et cognitive dont on présente certains aspects a donc pour objectif :

\begin{itemize}
\item  d’une part, d’élaborer une méthodologie précise de repérage et d’analyse des troubles de la communication compte tenu des caractéristiques de leur contexte naturel d’expression – "l’interaction verbale" – et compte tenu des variations de leur intensité ; cette méthodologie initiale est dite pragmatico-dialogique.
 \item d’autre part, d’enrichir la formalisation de manière à accéder à une représentation théorique de ces troubles de la communication en termes de "troubles de la pensée", en tant que ces troubles de la pensée correspondent à des faits psychologiques, en l’occurrence des processus mentaux qui spécifient le dysfonctionnement aux plans syntaxique, représentationnel et inférentiel ; cette seconde étape méthodologique est dite sémantico-formelle.
\end{itemize}

Le cadre formel utilisé est celui de la S-DRT \cite{asherlascaride03}. Il s’agit de combiner deux niveaux d’analyse pour pouvoir rendre compte des processus interprétatifs à l’œuvre dans les conversations : l’analyse du contenu sémantique et l’analyse pragmatique conversationnelle. La première est prise en charge par les Structures de représentation de discours segmenté (S-DRS) inspirées des DRS de la DRT, qui sont une construction syntaxique mise à jour au fil du flux conversationnel \cite{KampReyle93}. La conversation suppose en outre des relations pragmatiques entre actes de langage des interlocuteurs, dont la complexité donne lieu à une structuration hiérarchique appréhendée dès les années 1980 en linguistique \cite{Roulet85}. Nous proposons ici de formaliser ces relations à l’aide des relations rhétoriques de la S-DRT. Une conversation est alors interprétée par une double construction : celle d’un arbre hiérarchique reliant les actes, et celle des DRS représentant le contenu sémantique des actes. L’hypothèse que nous faisons est que l’interlocuteur schizophrène ne se conforme pas toujours aux règles devant prévaloir lors de cette double construction, ce qui explique le phénomène de rupture conversationnelle perçu par l’interlocuteur normal.
Dans la suite de cette présentation, nous nous focaliserons sur l'arbre hiérarchique, laissant hors de notre analyse la partie sémantique. 

\section{Présentation du corpus}

Pour sous-tendre ce travail, nous avons travaillé sur un corpus recueilli auprès de schizophrènes.
Nous revenons sur le protocole de définition de ce corpus.
Un total de 30 participants est inclus dans cette recherche, 18 hommes et 12 femmes (âge : 41,5 $\pm$ 16). Tous les sujets sont de langue maternelle française. 22 participants présentent un diagnostic de schizophrénie (SCH), 14 hommes et 8 femmes (âge : 45,0 $\pm$ 15.4). 8 participants ne présentent aucun diagnostic psychiatrique et ne prennent aucun traitement psychotrope ; ils constituent notre groupe contrôle (HC), il s’agit de 4 hommes et de 4 femmes (âge : 32,1 $\pm$ 14,3). Parmi les 22 schizophrènes, 15 d’entre eux prennent un traitement antipsychotique (SCH-A : moyenne en équivalent chlorpromazine en milligrammes par jour : 281 $\pm$ 118) et 7 ne prennent aucun traitement (SCH-S). La répartition en fonction des sous-types cliniques du DSM-IV\footnote{Manuel diagnostique et statistique des troubles mentaux, 4$^{e}$ édition, référence principale en psychologie et psychiatrie.} pour les patients schizophrènes est la suivante : 14 schizophrènes de type paranoïde (SCH-P) (dont 5 ne prennent aucun traitement antipsychotique) et 8 schizophrènes de type désorganisé (SCH-D) (dont 2 ne prennent aucun traitement antipsychotique). Les patients sont rencontrés dans deux structures hospitalières distinctes (Centres Hospitaliers Spécialisés de Troyes et de La Rochelle). Enfin, les 8 participants de notre groupe contrôle (HC) ont été rencontrés dans des lieux publics. Ils n’ont jamais présenté de pathologie psychiatrique et n’ont fait aucun usage de médicaments psychotropes par le passé.\\

\textbf{Le contexte des conversations.}
Les 30 entretiens sont réalisés par un interlocuteur psychologue-chercheur confronté soit à un interlocuteur diagnostiqué schizophrène, soit à un interlocuteur ne présentant pas de diagnostic psychiatrique. L'interaction est autant que possible non dirigée, la thématique concerne globalement  les occupations et/ou les préoccupations des interviewés. Tous les entretiens ont été retranscrits. L’analyse exhaustive de l’ensemble de notre corpus d’investigations empiriques nous a permis d’extraire 403 séquences conversationnelles (ou transactions\footnote{une transaction conversationnelle correspond à un nombre variable de tours de parole qui déploient une même thématique}). La table \ref{tab1} ci-dessous rend également compte de cette répartition.

Pour ce qui est des variables sociodémographiques, les comparaisons entre nos trois populations (SCH-P, SCH-D, HC) n’indiquent aucune différence significative en ce qui concerne le nombre d’années d’études (Fcal = 1,813 ; p = 0,189), l’âge (Fcal = 2,060 ; p = 0,147) ou le sexe ($\chi^2$ corrigé = 0,454 ; p = 0,80). Les facteurs "âge", "sexe" et "niveau d’éducation" n’interfèrent donc pas avec les résultats. La comparaison des deux populations de schizophrènes (SCH-P et SCH-D) quant aux traitements neuroleptiques auxquels elles sont soumises (moyenne en équivalent chlorpromazine en milligrammes par jour) ne montre aucune différence significative (Fcal = 0,113 ; p = 0,740). Chacune de ces deux populations comporte des patients traités par neuroleptiques et des patients non traités. La comparaison entre les populations schizophrènes de type paranoïde avec traitement (SCH-P-A) et schizophrènes de type désorganisé avec traitement (SCH-D-A), ne montre pas non plus de différence significative (Fcal = 0,588 ; p = 0,711) (Table \ref{tab1}). Le facteur "médication" n’interfère donc pas avec les résultats.

\newpage

\begin{table}[h]
\center
\begin{tabular}{l|c|c|c|c}
\cline{2-4}
&\textbf{schiz. paranoïdes}&\textbf{schiz. désorganisés}& \textbf{groupe contrôle}\\
\cline{2-4}
&SCH-P (n=14)&SCH-D (n=8)&HC (n=8)&\\
&(SCH-P-A / SCH-P-S)&(SCH-D-A / SCH-D-S)&&\\
&M$\pm$DS&M$\pm$DS&M$\pm$DS&\\
\cline{1-4}
\multicolumn{1}{|l|}{Sexe (H-F)}&	(10-4)&	(4-4)	&(4-4)&\\
\cline{1-4}
\multicolumn{1}{|l|}{Age en années} & 45,6 $\pm$ 17,0&	43,9 $\pm$ 13,4&	32,1 $\pm$ 14,3&\\
\cline{1-4}
\multicolumn{1}{|p{4cm}|}{Education (nombre d’années} &\multirow{2}{*}{9,4 $\pm$ 1,7}&\multirow{2}{*}{7,3 $\pm$ 3,5}	&\multirow{2}{*}{9,1 $\pm$ 1,4}&\\
\multicolumn{1}{|p{4cm}|}{d’études à compter du CP)}&&&\\
\cline{1-4}
\multicolumn{1}{|p{4cm}|}{Equivalent Chlorpromazine}
&173 $\pm$ 164 & 200 $\pm$ 183 & \multirow{2}{*}{0}&\\
\multicolumn{1}{|p{4cm}|}{en mg par jour}&(281 $\pm$ 107 / 0)	&(280 $\pm$ 148 / 0)&\\
\cline{1-5}
\multicolumn{1}{|l|}{\multirow{2}{*}{Nombre d’entretiens}} & 14 & 8&	\multirow{2}{*}{8}& \multicolumn{1}{|c|}{\multirow{2}{*}{30}}\\
\multicolumn{1}{|l|}{}&(9 / 5) & (6 / 2) &&\multicolumn{1}{|c|}{}\\
\hline
\multicolumn{1}{|l|}{Nombre de transactions} & 208 & 150 & \multirow{2}{*}{45}& \multicolumn{1}{|c|}{\multirow{2}{*}{403}}\\
\multicolumn{1}{|l|}{conversationnelles} & (146 / 62)	& (108 / 42) & &\multicolumn{1}{|c|}{}\\
\hline
\end{tabular}

n : nombre de sujets ; 
M : moyenne ; 
DS : déviation standard.

\caption{Caractéristiques biographiques de la population étudiée et caractéristiques du corpus d’investigations empiriques.\label{tab1}}
\end{table}

\textbf{Ruptures}. La répartition des séquences conversationnelles "discontinues" ou "non discontinues" varie selon la population à laquelle appartient l’interlocuteur.
Les deux populations de schizophrènes produisent des ruptures pragmatiques et conversationnelles comme le montre la table \ref{tab2} contrairement aux sujets de contrôle.
Par ailleurs, les deux populations de schizophrènes (paranoïdes vs désorganisés) ne sont pas significativement distinctes relativement à la production de ruptures non décisives.
Lorsque les schizophrènes ne sont soumis à aucun traitement (SCH-S), on constate plus de discontinuités non décisives chez les schizophrènes désorganisés (SCH-D) que chez les schizophrènes paranoïdes (SCH-P) ($\chi^2$ de croisement = 22,015, p < 0,001). Par contre, lorsqu’ils sont soumis à un traitement antipsychotique (SCH-A), il y a plus de discontinuités non décisives chez les paranoïdes (SCH-P) que chez les désorganisés (SCH-D) ($\chi^2$-deux de croisement = 13,141, p < 0,001).
(cf table \ref{tab3})

\begin{table}[h]

\begin{tabular}{c|c|c|c|c|c|c|c|}
\cline{2-8}

\multirow{3}{*}{}&\multirow{3}{*}{\textbf{SCH-P}}&\multirow{3}{*}{\textbf{SCH-D}}& \multirow{3}{*}{\textbf{HC}}& \multicolumn{4}{c|}{\textbf{p-values}}\\
\cline{5-8}
&&&&SCH&	SCH-P&	SCH-D	&SCH-P \\
&&&&vs HC& vs HC& vs HC&vs SCH-D\\
\hline
\multicolumn{1}{|c|}{Séquences non discontinues}      &128 (62\%)	&100 (67\%)	&44 (98\%)&	<.001&	<.001&	<.001&	.319\\
\cline{1-4}
\multicolumn{1}{|c|}{Séquences avec discontinuité}	& 80 (38\%)&	50 (33\%)&	1 (2\%)
&&&&\\
\cline{1-4}
\hline
\hline
\multicolumn{1}{|c|}{- Séquences avec discontinuité non décisive}&	71 (34\%)	&50 (33\%)&	1 (2\%)&	<.001& <.001 &<.001&.649\\
\cline{1-4}
\cline{1-4}
\multicolumn{1}{|c|}{- Séquences avec discontinuités décisives}&9 (4\%)&0&0&&&&\\
\hline
\multicolumn{1}{|c|}{Total}&             	208	&150&	45&&&&\\
\hline
\end{tabular}

\caption{Présence ou non de discontinuités en fonction de la population à laquelle l’interlocuteur appartient. \label{tab2}}
\end{table}

\begin{table}[h]
\center
\begin{tabular}{cl|c|c|c|}
\cline{3-4}
&&\textbf{SCH-P}&\textbf{SCH-D}\\
\hline
\multicolumn{1}{|c|}{\multirow{4}{*}{SCH-S}}&Séquences avec discontinuité non décisive&12 (20\%)&28 (67\%)\\
\cline{2-4}
\multicolumn{1}{|c|}{}&Séquences non discontinues (33\%)&\multirow{1}{*}{47 (80\%)}&\multirow{1}{*}{14}\\
\cline{2-4}
\multicolumn{1}{|c|}{}&Total&59&42\\
\hline
\hline
\multicolumn{1}{|c|}{\multirow{4}{*}{SCH-A}}&Séquences avec discontinuité non décisive&59 (42\%)&22 (20\%)\\
\cline{2-4}
\multicolumn{1}{|c|}{}&Séquences non discontinues (80\%)&\multirow{1}{*}{81 (58\%)}&\multirow{1}{*}{86}\\
\cline{2-4}
\multicolumn{1}{|c|}{}&Total&140&108\\
\hline

\end{tabular}
\caption{Présence de discontinuités non décisives en fonction de la forme clinique de l’interlocuteur-patient et  de la médication. \label{tab3}}
\end{table}

\textbf{Répartition des séquences conversationnelles présentant ou non une discontinuité décisive selon la population à laquelle appartient l’interlocuteur.}
Seul le sous-corpus des schizophrènes paranoïdes (SCH-P) présente des séquences à discontinuité décisive. Cette population se distingue donc significativement de la population des schizophrènes désorganisés (SCH-D) (test binomial, p = 0,002) ou des sujets ne présentant aucun diagnostic psychiatrique – HC – (test binomial, p = 0,002). Parmi les neufs séquences à discontinuité décisive mises au jour, trois sont issues du sous-corpus "schizophrènes paranoïdes sans traitement" (SCH-P-S) et les six autres sont issues du sous-corpus "schizophrènes paranoïdes avec traitement" (SCH-P-A).
Il est donc raisonnable de focaliser l'étude sur les discontinuités décisives des schizophrènes paranoïdes dans la suite de la formalisation. 

\textbf{Le choix des extraits.}
 L’analyse des transactions conversationnelles au sein desquelles on reconnaît des ruptures non décisives oblige le plus souvent à une interprétation pragmatique faisant intervenir le contexte conversationnel ou situationnel d’une manière telle que le patient schizophrène pourrait très bien n’être considéré comme responsable du caractère incohérent de l’interaction verbale qu’en termes d’infraction à la coopération. Ces discontinuités affectent des relations entre actes continus au niveau intra-intervention ou bien des relations inter-interventions au sein d’échanges confirmatifs.
En l’état actuel du développement de notre modèle, deux sortes de ruptures peuvent être considérées comme décisives. La première progresse au fil de transactions conversationnelles qui prennent la forme d’un échange. La distribution des actions (actes de langage) et la progression argumentative s’accomplissent de manière symétrique. Nous avons qualifié ce type de discontinuité de débrayage conversationnel \cite{MusiolTrognon96}. Le second type de discontinuité décisive affecte la rationalité de l’intervention complexe ; il s’agit d’un cas de défectuosité de l’initiative conversationnelle. Les séquences conversationnelles à ruptures décisives ne résistent pas à l’épreuve du principe logique de non-contradiction alors que les séquences conversationnelles à ruptures non décisives présentent des caractéristiques incongrues ou des formes d’incohérences étayant des infractions comportementales de type "normatives". Les discontinuités de type décisives affectent l’organisation et la progression du cours de transactions conversationnelles qui comptent toujours au moins trois tours de parole et la plupart du temps beaucoup plus. Elles impliquent trois constituants à chaque fois, c’est-à-dire trois constituants de rang "intervention" dans le cas du débrayage conversationnel ou trois constituants de rang "acte" dans le cas de la défectuosité de l’initiative conversationnelle \cite{Musiol2009}. 
 
\section{Présentation du formalisme}
La formalisation des conversations est réduite aux éléments pertinents pour notre analyse, ce qui signifie que nous délaissons ce qui ne paraît pas devoir jouer de rôle explicatif dans les ruptures. Ainsi la représentation du contenu sémantique est-elle réduite au minimum, à savoir le thème conversationnel. Cela est justifié par le fait que nombre de ruptures prennent corps autour d’expressions sous-spécifiées dont la valeur est paramétrée, le plus souvent par le thème contextuel. Dans une représentation complète, l’indice thématique pourrait figurer parmi les marqueurs syntaxiques de l’univers de la S-DRS. Cet indice étant constant sur une succession d’échanges, nous pouvons regrouper les actes en classes selon le thème.

Les relations sont de différents types. A côté des relations usuelles de la S-DRT (narration, élaboration), les conversations étudiées nous ont conduits à intégrer des relations d’autres types incluant celles correspondant aux ajustements méta-conversationnels.
Comme dans la S-DRT, les relations sont coordonnantes (comme la narration, la réponse à une question, etc.) ou subordonnantes (comme l’explication, la requête de clarification, etc.). Nous récapitulons les types de relation utilisés dans le tableau suivant :
\begin{center}
\begin{tabular}{c|l|l|}
\cline{2-3}
&Subordonnantes& 	Coordonnantes \\
\hline
\multicolumn{1}{|c|}{\multirow{3}{*}{Conversationnelles}} & Question & Réponse\\

\multicolumn{1}{|c|}{}&Elaboration&Narration \\

\multicolumn{1}{|c|}{}&Contre-élaboration&\\
\hline
\multicolumn{1}{|c|}{\multirow{3}{*}{Méta-conversationnelles}} & Requête de clarification & Clarification\\

\multicolumn{1}{|c|}{}& Conduite & C-Réponse\\
\multicolumn{1}{|c|}{}& Phatique &\\

 \hline
\end{tabular}
\end{center}

Pour l’analyse des conversations pathologiques, nous proposons systématiquement la construction simultanée de deux représentations, une par interlocuteur. 
Du côté du sujet schizophrène, nous faisons un postulat de rationalité analogue au principe de charité \cite{Quine60} ce qui signifie que la représentation est dépourvue de contradictions au niveau sémantique ; s’il y a ruptures, elles interviennent donc au niveau pragmatique, par l’infraction de règles de construction des arbres de la S-DRT. La situation n’est pas la même en face. Dans les conversations qui constituent les corpus étudiés, l’interlocuteur normal est un psychologue en charge de poursuivre l’entretien : il fait ainsi en sorte de "réparer" la structure conversationnelle après une rupture, même si cette dernière est telle qu’elle aurait provoqué l’interruption d’une conversation analogue dans une autre situation. On a alors un postulat correspondant à cette prescription, à savoir la construction d’une représentation respectant les contraintes pragmatiques ; cette option entraîne l’apparition d’inconsistances au niveau sémantique. La dualité des représentations conversationnelles exprime une dualité de points de vue sur la conversation : le sujet schizophrène est apparemment contradictoire à l'épreuve du comportement dialogique du sujet normal, si bien que la conversation fonctionne mais que la représentation du monde co-construite est inconsistante ; à l’inverse, le dysfonctionnement du sujet schizophrène résidant dans sa gestion des relations pragmatiques, la représentation du monde construite par la conversation ne souffre pas de ce défaut.

Le traitement des extraits du corpus sont des dialogues dont l'un des participants n'a d'autres intentions que de continuer l'échange. Nous considérons donc ces échanges comme un discours, porté par le schizophrène. Les interventions de l'interlocuteur non-schizophrène sont alors souvent des relances ou des phatiques, d'où l'arborescence particulière des représentations. Nous nous attachons à suivre les structures usuelles de la S-DRT. Cependant, comme nous nous attachons uniquement à la structure pragmatique de la représentation, nous évacuons toutes représentations sémantiques. Seule la notion de thème est conservée. Pour marquer la cohérence thématique dans l'échange, nous incluons les items dans une boîte en pointillés.

Nous assumons que les échanges sont des extraits d'échanges plus larges, dont le point de départ est un noeud particulier à sémantique vide. Ainsi, quel que soit le traitement choisi, il est toujours possible de rattacher un élément, au pire à cette racine. L'analyse des extraits nous conduit à mettre en avant deux transgressions de l'usage standard de la S-DRT: d'une part des ruptures de la frontière droite (rattachement à l'intérieur de la structure alors que la S-DRT suppose qu'ils se font sur la frontière droite) et d'autre part, des montées au travers de la structure sans clôture basse acceptable (inconsistance de la représentation). Il est en effet fréquent dans le corpus d'identifier des items qui servent à clore une partie de l'échange et à ouvrir un nouveau thème.

De manière globale, sur le corpus, nous avons identifié trois ruptures de la frontière droite et cinq montées sans complétude de la sous-structure.
Nous présentons l'analyse de deux exemples présentant les deux types de discontinuités.
Dans le premier, le schizophrène change deux fois de thème passant d'une interprétation de la mort symbolique dans le cadre de la politique, et la mort usuelle. Il est clair que ces deux interprétations de la mort sont directement reliées, mais elles expriment deux réalités différentes.

\begin{figure}[htd]
\begin{minipage}{0.5\textwidth}
\begin{verse}
B$_{124}$ : Oh ouais ($\uparrow$) et pis compliqué ($\downarrow$) et c’est vraiment très très compliqué ($\rightarrow$) la politique c’est quelque chose quand on s’en occupe faut être gagnant parce qu’autrement quand on est perdant c’est fini quoi ($\downarrow$) 

A$_{125}$ : oui

B$_{126}$ : J. C. D. est mort, L. est mort, P. est mort euh (...) 

A$_{127}$ : Ils sont morts parce qu’ils ont perdu à votre avis ($\uparrow$)

B$_{128}$ : Non ils gagnaient mais si ils sont morts, c’est la maladie quoi c’est c’est ($\rightarrow$)

A$_{129}$ : Ouais c’est parce qu’ils étaient malades, c’est pas parce qu’ils faisaient 
de la politique ($\uparrow$)

B$_{130}$ : Si enfin ($\rightarrow$)

A$_{131}$ : Si vous pensez que c’est parce qu’ils faisaient de la politique ($\uparrow$)

B$_{132}$ : Oui tiens oui il y a aussi C. qui a accompli un meurtre là ($\rightarrow$) il était présent lui aussi qui est à B. mais enfin ($\rightarrow$) c’est encore à cause de la politique ça
\end{verse}
\end{minipage}
\begin{minipage}{0.5\textwidth}
\begin{verse}
G$_{82}$ :	(...) l’an dernier euh ($\rightarrow$) j’savais pas comment faire j’étais perdue et pourtant j’avais pris mes médicaments j’suis dans un état 	vous voyez même ma bouche elle est sèche j’suis dans un triste état

V$_{83}$ :	Vous êtes quand même bien ($\uparrow$)

G$_{84}$ :	J’pense que ma tête est bien mais on croirait à moitié ($\downarrow$) la moitié qui va et la moitié qui va pas j’ai l’impression de ça vous voyez ($\uparrow$)

V$_{85}$ :	D’accord

G$_{86}$ :	Ou alors c’est la conscience peut être la conscience est ce que c’est ça ($\uparrow$)

V$_{87}$ :	Vous savez ça arrive à tout le monde d’avoir des moments biens et des moments où on est perdu

G$_{88}$ : 	Oui j’ai peur de perdre tout le monde

V$_{89}$ :	Mais ils vont plutôt bien vos enfants ($\uparrow$)

G$_{90}$ :	Ils ont l’air ils ont l’air mais ils ont des allergies ils ont ($\rightarrow$) mon petit fils il s’est cassé le bras à l’école tout ça 

V$_{91}$ :	C’est des petits incidents de la vie quotidienne vous savez ($\uparrow$)

G$_{92}$ :	Oui oui
\end{verse}
\end{minipage}
\caption{Deux extraits du corpus\label{discours}.}
\end{figure}

\begin{figure}[htd]
\center
\includegraphics[height = 6.1 cm]{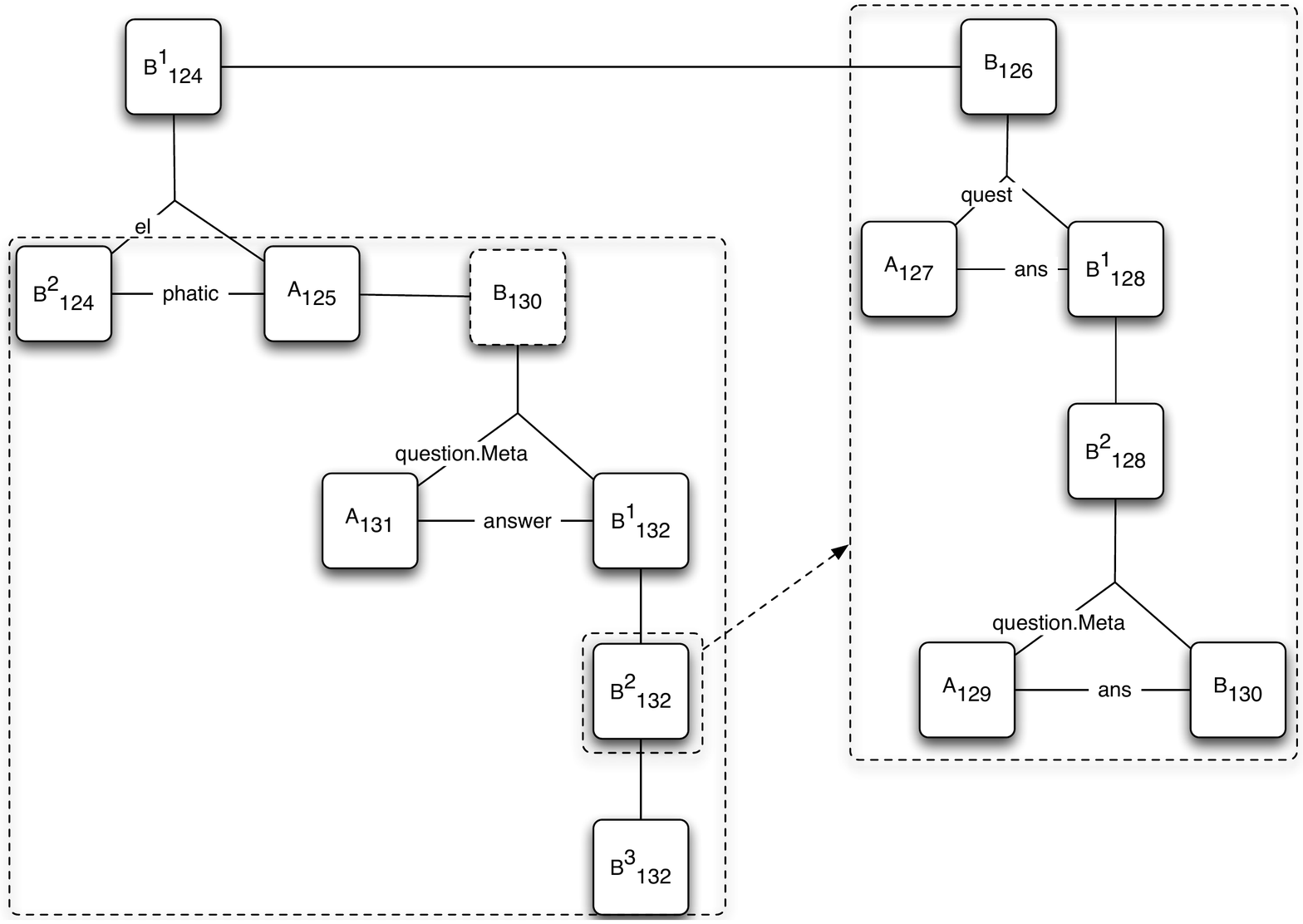}
\includegraphics[height = 6.1 cm]{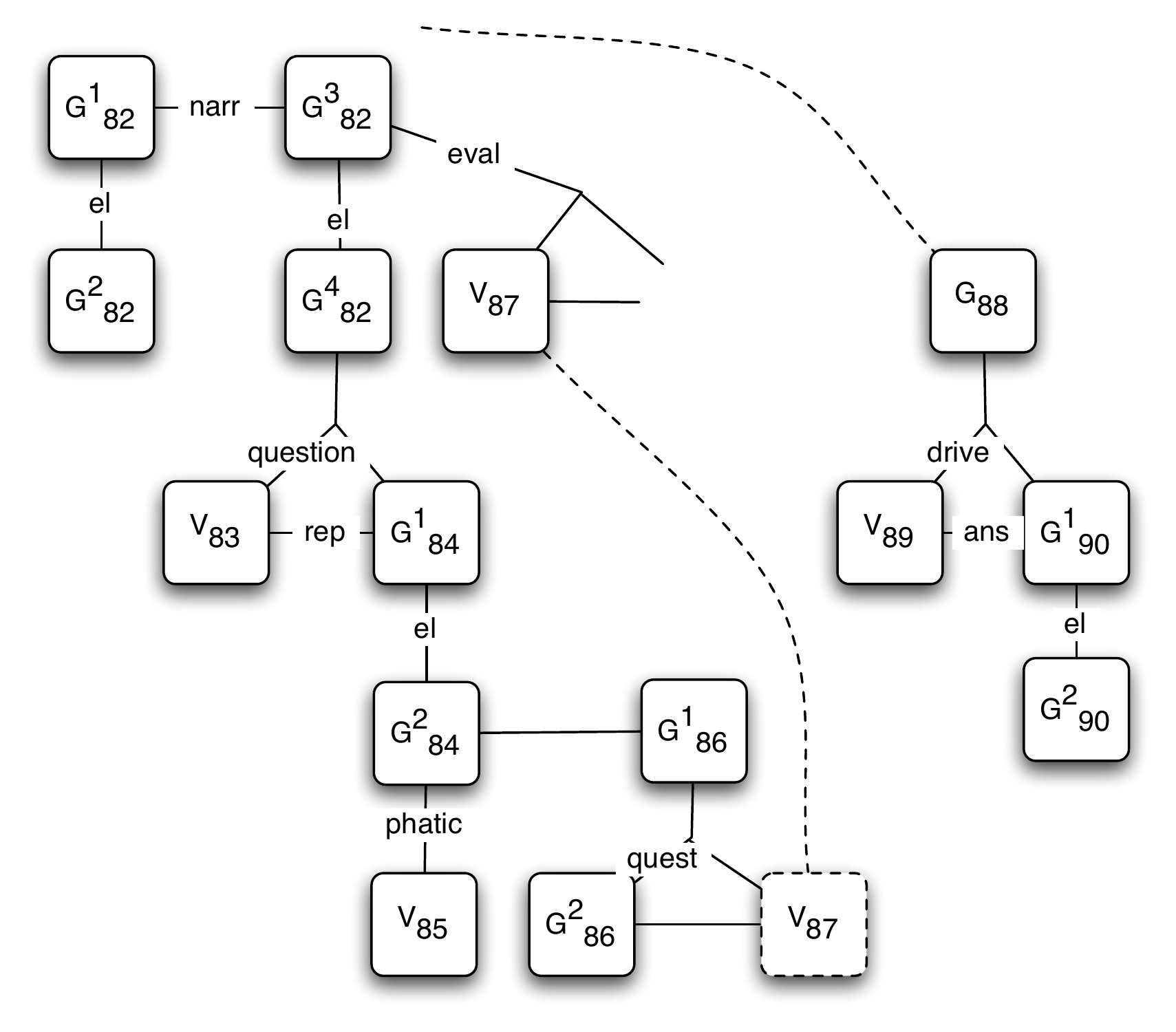}
\caption{Représentations sous forme d'arbre de S-DRS des deux discours\label{repdialogue}.}
\vspace{-10ex}
\end{figure}

La représentation de cet extrait est donnée dans la figure \ref{repdialogue}.
L'intervention B$_{126}$ ouvre une nouvelle boîte thématique passant de la mort symbolique à la mort réelle sans poser de problème, la première partie de la dérivation ayant été correctement close.
La dérivation se poursuit jusqu'à l'intervention B$_{130}$ qui est ambiguë et dont le contenu sémantique est faible. Quel que soit le choix fait, pour le rattachement, l'intervention B$_{132}$ résout l'ambiguïté.
Et c'est là que se pose un premier problème. La contrainte de la frontière droite sur cette structure ne permet pas le rattachement de B$_{130}$ sur A$_{125}$, pourtant, c'est en ce point que B$_{130}$ doit thématiquement être rattachée.
Nous avons une première discontinuité.
Un problème similaire se pose sur l'intervention B$^3_{132}$ qui cette fois doit arbitrairement être rattachée sur la seconde boîte thématique, ce qui n'est pas possible.

Dans le second arbre de la figure \ref{repdialogue}, la patient utilise l'ambiguïté sur l'interprétation de la perte.
Le psychologue donne un exemple de montée au travers de la structure en V$_{87}$, ouvrant une nouvelle partie de l'échange.
L'intervention en G$_{88}$, la réponse du patient, est alors une montée dans la dérivation, mais sans fermeture de la sous-dérivation. C'est le second type de problème identifié dans ces échanges. 

\section{Discussion}

Nous avons discuté plus haut de la nécessité de construire deux représentations conversationnelles, une pour le sujet schizophrène, l’autre pour le sujet normal. La dualité des représentations conversationnelles se traduit par une dualité des représentations sémantiques (ou représentations du monde construite dans la conversation) : contradictoire chez le sujet normal, elle est en revanche parfaitement consistante pour le sujet schizophrène. 

Il y a donc du jeu dans la localisation de l’inconsistance chez les sujets schizophrènes. Nous aurions pu choisir de calquer la représentation conversationnelle du schizophrène sur celle du sujet normal. Cela aurait cependant produit une continuité entre les deux interlocuteurs, là où nous pensons qu’il n’y en a pas : ce qui se manifeste pour le sujet normal, qui relève de la pathologie du schizophrène, a toutes les chances de se manifester différemment pour ce dernier. Nous avons donc fait le choix de postuler que le sujet schizophrène est rationnel au sens le plus commun du terme, est qu’il se conforme aux normes de la logique tout autant que le sujet normal : il n’y a pas plus de contradictions dans la représentation du monde d’un schizophrène que dans celle d’un sujet normal – même si du point de vue de ce dernier, il y a des contradictions (ce qui rejoint certains traits symptomatiques habituels de la pathologie). L’inconsistance est localisée dans la hiérarchie pragmatique, qui relève de l’interaction entre les deux sujets en conversation. Fondamentalement c'est un déficit interactionnel, plutôt qu’un dysfonctionnement rationnel, qui est ici en jeu.

La suite de ce projet de recherche est d'augmenter le nombre d'échanges dans le corpus. Actuellement, les échanges sont en cours de recueillement et de transcription. Ces nouveaux éléments permettront de valider l'hypothèse de départ quant à l'utilisation non usuelle de la pragmatique du discours du groupe de schizophrènes paranoïdes. Enfin, la validation de ces travaux expérimentaux valide réciproquement l'adéquation des structures produites par la S-DRT aux modèles mis en oeuvre dans les échanges standards.

\vspace{-1ex}

\bibliographystyle{taln2002}
\bibliography{biblio}

\end{document}